\documentclass{article}
\usepackage{spconf,amsmath,graphicx}
\usepackage{todonotes}
\usepackage{booktabs}
\usepackage{subfigure}
\usepackage{romannum}
\usepackage{graphicx}
\usepackage{soul}

\usepackage[font=small,labelfont=bf]{caption}

\usepackage{lipsum}

\newcommand\blfootnote[1]{%
  \begingroup
  \renewcommand\thefootnote{}\footnote{#1}%
  \addtocounter{footnote}{-1}%
  \endgroup
}

\title{Fast-converging Conditional Generative Adversarial Networks for Image Synthesis} 

\name{Chengcheng Li, Zi Wang, Hairong Qi 
}
\address{Department of Electrical Engineering and Computer Science\\ University of Tennessee, Knoxville, TN 37996}
\begin{document}
\maketitle
\begin{abstract}
Building on top of the success of generative adversarial networks (GANs), conditional GANs attempt to better direct the data generation process by conditioning with certain additional information. Inspired by the most recent AC-GAN, in this paper we propose a fast-converging conditional GAN (FC-GAN). In addition to the real/fake classifier used in vanilla GANs, our discriminator has an advanced auxiliary classifier which distinguishes each real class from an extra `fake' class. The `fake' class avoids mixing generated data with real data, which can potentially confuse the classification of real data as AC-GAN does, and makes the advanced auxiliary classifier behave as another real/fake classifier. As a result, FC-GAN can accelerate the process of differentiation of all classes, thus boost the convergence speed. Experimental results on image synthesis demonstrate our model is competitive in the quality of images generated while achieving a faster convergence rate.\blfootnote{Accepted to be published in: Proceedings of the 2018 IEEE International Conference on Image Processing. Personal use of this material is permitted. However, permission to reprint/republish this material for advertising or promotional purposes or for creating new collective works for resale or redistribution to servers or lists, or to reuse any copyrighted component of this work in other works must be obtained from the IEEE.}
\end{abstract}
\section{Introduction}
\label{sec:intro}

Generating high-resolution and photo-realistic images has always been one of the long-standing goals in the generative modeling community. Image synthesis is of significance to many applications, such as image editing, image inpainting, image translation, pattern recognition, etc. \cite{brock2016neural,yeh2017semantic,zhang2017age,isola2017image,li2015iris}. In recent years, deep generative models have brought breakthroughs in this area. Three main branches of methods have been developed, including Variational Auto-Encoder (VAE) \cite{doersch2016tutorial}, Generative Adversarial Networks (GANs) \cite{goodfellow2014generative}, and PixelRNN/PixelCNN \cite{oord2016pixel,van2016conditional}. 

Among these methods, VAEs tend to blur the rich details in the generated images and the sequential generation of PixelRNN/PixelCNN is time-consuming. On the contrary, GANs can quickly generate images with more photo-realistic details. A GAN model consists of two competing players: the discriminator and the generator, where the generator takes as inputs latent variables and generates synthesized data to fool the discriminator while the discriminator tries to distinguish between synthesized data and real ones.  

However, the vanilla GANs have no control over the mode of the generated results. For instance, if we train a vanilla GAN model on a digits dataset containing digits $0, 1, ..., 9$, then with a latent variable as input, a random digit among these $10$ digits will be generated. In many cases, it is necessary to direct the generation process with certain conditions, such as age conditioning for face regression/progression, text conditioning for text-to-image translation, and image conditioning for image-to-image translation \cite{zhang2017age,isola2017image,isola2016image,reed2016learning,ma2017pose}.  

There have been a couple of studies focusing on different ways of conditioning the vanilla GANs \cite{mirza2014conditional,kwak2016ways,odena2016conditional,bodla2018semi}. For example, the pioneer, CGAN \cite{mirza2014conditional} performs the conditioning by feeding conditioned attributes into both the generator and discriminator as  additional inputs while keeping the other parts the same as vanilla GANs. Since CGAN does not have any specific constraint on the classes of the generated data, it can easily neglect the conditioned attributes without deliberately-designed architectures. The most recent development in the family of conditional GAN is AC-GAN \cite{odena2016conditional}, which introduces an auxiliary classifier for the discriminator. 
The auxiliary classifier assigns each real sample to its specific class and each generated sample to the class corresponding to the generator input. The overall loss is then defined by combining the discrimination loss (source loss) between real/fake samples and the classification loss over all conditioned classes. The auxiliary classifier better directs AC-GAN to generate desired images of different classes. However, assigning fake data with their real class labels the same way as real data can potentially confuse the auxiliary classifier.

Inspired by AC-GAN, we propose FC-GAN that introduces an advanced auxiliary classifier for the purposes of fast convergence and improved quality. It achieves these goals in two aspects. First, the auxiliary classier distinguishes each attribute class from an extra `fake' class. In this way, real data are categorized into real classes and generated data are categorized into a `fake' class rather than having both real and generated samples categorized into real classes as AC-GAN does.  The proposed advanced auxiliary classifier effectively  accelerates the process of differentiation of each class. Second, the existence of the `fake' class makes the auxiliary classifier also behave like another real/fake classifier, which can potentially boost the convergence speed.

The paper is organized as follows. Section \ref{sec:approach} demonstrates the proposed FC-GAN in detail. The comprehensive experimental results are shown in Section \ref{sec:exp}. The work is concluded in Section \ref{sec:conclusion}.
\section{Proposed Approach}
\label{sec:approach}
\subsection{GAN}
A GAN model consists of two competing players: the discriminator $D(\theta_d)$, and the generator $G(\theta_g)$. The generator and discriminator have opposite objectives during training, where the discriminator is trained toward distinguishing between synthesized and real data while the generator is trained to fool the discriminator with synthesized data.

The objective function for GAN can be formulated as a minimax optimization problem in Eq. (\ref{eq:1}), 
\begin{multline}
\min \limits_{G} \max \limits_{D} V(D,G) = 
E_{x \sim p_{data}(x)}[\log(D(x))] + \\
E_{z \sim p_{z}(z)}[\log(1-D(G(z)))]
\label{eq:1}
\end{multline}
where $p_{data}(x)$ denotes the true distribution of real data $X_{real}$, and $p_z(z)$ is the prior distribution of latent variable $z$, also known as noise. The generator $G$ takes as input samples $z$ from $p_z$ and outputs synthesized data $X_{fake} = G(z)$. The discriminator is a real/fake classifier which distinguishes synthesized data from real ones. 
According to game theory, in the space of arbitrary functions for $G(\theta_d)$ and $D(\theta_g)$, a unique solution exists when the Nash equilibrium is achieved \cite{goodfellow2014generative}.

\subsection{CGAN and AC-GAN}
The basic GAN framework can be extended to a conditional GAN model with certain auxiliary information $y$. CGAN performs the conditioning by feeding $y$ into both the discriminator and the generator as an extra input. $y$ and the latent variable $z$ are combined as the input for the generator while $y$ and  the sample $x$ are concatenated as the input for the discriminator.

AC-GAN introduces an auxiliary classifier built on the discriminator to give a probability distribution over the class labels for both real data and generated data.
The generator takes both latent variables and class information as input to generate synthesized images $X_{fake}$. Every generated sample has a corresponding class label $c$ in addition to the latent variable $z$. The discriminator outputs two probability distributions. One is over sources, i.e., real or fake data, and the other is over the class labels, denoted as $P(S|X)$ and $P(C|X)$, respectively. The overall objective function combines the source loss and classification loss. 
\subsection{Proposed FC-GAN}
\begin{figure}[htb]
\centering
\begin{minipage}{0.9\columnwidth}
\centering
\includegraphics[width=\textwidth]{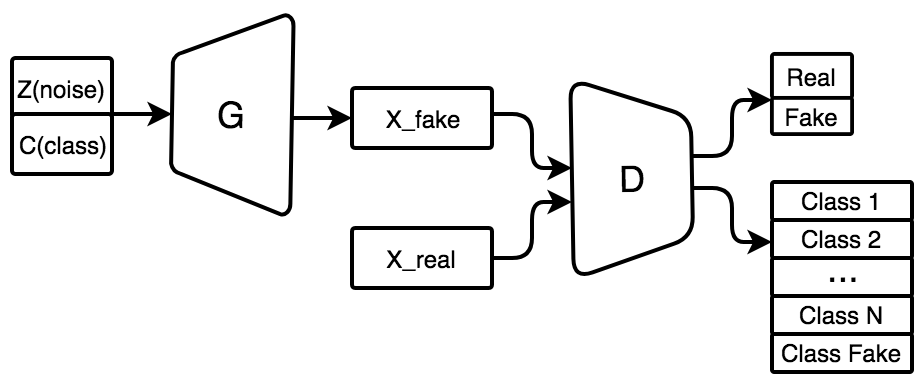}
\end{minipage}

\caption{The FC-GAN framework.}
\label{fig:FC-GAN-framework}
\end{figure}

The proposed FC-GAN, shown in Fig.~\ref{fig:FC-GAN-framework}, belongs to the family of conditional generative adversarial networks. The setting here is the same as other conditional GANs, where the real data $X_{real}$ could be categorized into $N$ classes, $[C_1, ..., C_N]$ according to the conditioned information $y$. We introduce a new class $C_{fake}$ to denote the category of data generated by the generator. We thus design an advanced auxiliary classifier on  top of the discriminator, which gives a probability over $N+1$ class labels, corresponding to $[C_1, ..., C_N, C_{fake}]$. 
{Unlike AC-GAN, generated data are not assigned to one of the $N$ real classes during the training of the discriminator.} 

We perform the conditioning on the generator $G$ by feeding class information $C$ as well as latent variable $Z$ as the input. Like AC-GAN, we also define two losses, the source loss and the classification loss. However, because of the additional $C_{fake}$ class, the classification loss is defined differently.

For the source loss, we train the discriminator $D$ to maximize the probability of assigning real data and generated data to their correct source classes $S$, i.e., real vs. fake.
The source loss function for the discriminator $D$ is defined as 
\[
\begin{array}{lll}
L_s^{D}&=& -(E[\log P(S=real|X_{real})] + \\
& &E[\log P(S=fake|X_{fake})]).
\end{array}
\] 
And we train the generator $G$ to maximize the probability that $D$ assigns the generated data to real data class, with the loss function defined as 
\[
L_s^{G}=-E[\log P(S=real|X_{fake})].
\]

For the classification loss, we train the discriminator $D$ to maximize the probability of assigning real data to the correct class $C_i$ out of $[C_1, ..., C_N]$ and maximize the probability of assigning generated data to the class $C_{fake}$, i.e., the classification loss for $D$ is defined as
\[
\begin{array}{lll}
L_c^{D} & = & -(E[\log P(C = C_i|X_{real})] +\\
& & E[\log P(C = C_{fake}|X_{fake})]).
\end{array}
\] 
{And $G$ is trained to maximize the probability that $D$ assigns each generated sample to the class $C_i$ corresponding to the input class of $G$, i.e., the classification loss for $G$ is defined as} 
\[
L_c^{G}=-E[\log P(C = C_i|X_{fake})].
\]

The overall loss function consists of both the source loss and the classification loss, where  the overall loss for the discriminator $D$ is defined in Eq. (\ref{eq:2}),
\begin{equation}
  \renewcommand{\arraystretch}{1.5}%
  \resizebox{!}{14.5mm}{
    $\begin{array}{l}
     L^{D} = L_s^{D} + L_c^{D}\\
     =-(E[\log P(S=real|X_{real})] +E[\log P(S=fake|X_{fake})])-\\(E[\log P(C = C_i|X_{real})] +E[\log P(C = C_{fake}|X_{fake})])\\
=-(E_{x \sim p_{data}(x)}[\log{D_1(x|y)}]+E_{z \sim p_{z}(z)}[\log{(1-D_1(G(z|y)))}])  \\
     -( E_{x \sim p_{data}(x)}[\log D_2^i(x_i|y)] + E_{z \sim p_{z}(z)}[\log (1-D_2^{fake}(G(z|y)))])
    \end{array}$%
  }
  \label{eq:2}
  \end{equation}
and the overall loss for the generator $G$ is defined in Eq. (\ref{eq:3}),
\begin{equation}
  \renewcommand{\arraystretch}{1.5}%
  \resizebox{.45\textwidth}{!}{
    $\begin{array}{l}
     L^{G}= L_s^{G} + L_c^{G}\\
     =-E[\log P(S=real|X_{fake})]-E[\log P(C = C_i|X_{fake})]\\
     =-E_{z \sim p_{z}(z)}[\log{(D_1(G(z|y)))}])\\
     -E_{z \sim p_{z}(z)}[\log (D_2^{i}(G(z|y)))]),
    \end{array}$%
  }
  \label{eq:3}
  \end{equation}
where $y$ represents the conditioned information, $D_1$ and $D_2$ represent the real/fake classifier and the advanced auxiliary classifier, respectively, the subscript in $x_i$ denotes that the ground truth class of sample $x$ is $C_i$, $D^i_2(x_i)$ denotes the probability that the advanced auxiliary classifier assigns the sample $x_i$ to the class $C_i$, and $D^{fake}_2(\hat x)$ denotes the probability that the advanced auxiliary classifier assigns the generated sample $\hat{x}=G(z|y)$ to the class $C_{fake}$.

\section{Experiments and Results}
\label{sec:exp}
\subsection{Datasets and Network Structure}
We utilize two popular benchmark datasets, MNIST \cite{lecun1998mnist} and CIFAR-10 \cite{liu2015deep}, for the evaluation. 
The MNIST dataset contains 10 digit classes, and images are centered and resized to size  $28\times28 $. 
The CIFAR-10 dataset consists of $32\times32$ color images in 10 classes. For both datasets, we train the models conditioned on their class labels. 

We focus our comparisons to AC-GAN since it represents the state-of-the-art in the development of conditioning GAN. 
As far as we know, there are  no official public implementations for AC-GAN. We adopt the version included in Keras \cite{keras-acgane} and implement the proposed FC-GAN based on it. 

The FC-GAN structure for MNIST is shown in Table \ref{tab:fc-gan_architecture-mnist} where $Conv$, $Deconv$, and $FC$ represent the convolution layer, deconvolution layer, and fully connected layer, respectively. The structure for CIFAR-10 is similar with the only difference being the size of the feature maps due to the different input image sizes. The number of classes $N$ for both MNIST and CIFAR-10 datasets is $10$. Following  DCGAN \cite{radford2015unsupervised, salimans2016improved}, in the generator, we use ReLU activation for all hidden layers and Tanh for the output layer. We use Leaky ReLU activation for all hidden layers in the discriminator. 
The number of outputs of the auxiliary classifier is $N+1$, corresponding to the $N+1$ classes, $[C_1, ..., C_N, C_{fake}]$. 

We use the uniform distribution on $[-1,1]$ for the noise $Z$ with a dimension of $100$. We also experiment with Gaussian distribution but find no performance difference. The Adam optimizer \cite{kingma2014adam, wang2018deep} is employed with parameters $\alpha = 0.00002$, $ \beta_1= 0.5$, $\beta_2=0.999$. And the batch size is $100$. The weights are initialized with truncated normal distribution. The above settings are consistent for both MNIST and CIFAR-10 datasets.
\begin{table}[htb]
\centering
{\resizebox{0.9\linewidth}{!}{\begin{tabular}{ccc|ccc}
\hline
\hline
\multicolumn{3}{c}{Discriminator} & \multicolumn{3}{c}{Generator}\\
\hline
Layer & Filter/Stride & Output Size & Layer & Filter/Stride & Output Size\\ \\
Conv1  & $3\times3$ /2 & $32 \times 14 \times 14$ & FC1 & & 1024\\
Conv2  & $3\times3$ /1 & $64 \times 14 \times 14$ & FC2 & & $128\times 7 \times 7$\\
Conv3  & $3\times3$ /2 & $128 \times 7 \times 7$& Deconv1 & $5 \times 5 /2$& $256 \times 14 \times 14$\\
Conv4  & $3\times3$ /1 &$ 256 \times 7 \times 7$ & Deconv2& $5 \times 5 /2$& $128 \times 28 \times 28$\\
D\_source & & 1 &Conv3 & $2 \times 2 /1$& $1\times 28 \times 28$ \\
D\_class & & N+1 & & &\\
\hline
\hline
\end{tabular}}
\caption{The FC-GAN architecture for MNIST dataset.}
\label{tab:fc-gan_architecture-mnist}

}
\end{table}
\subsection{Qualitative and Quantitative Analyses}
Comparison between different generative models is very challenging and good performance with respect to one criterion does not imply good performance with respect to other criteria \cite{theis2015note}. We adopt best-known evaluation criteria, including visual fidelity, Parzen window, and inception score, to have a qualitative and quantitative analysis of the proposed model.
\subsubsection{Visual Fidelity}
\begin{figure}[htb]
\centering
\begin{minipage}{\columnwidth}
\centering
\subfigure[\scriptsize AC-GAN 10 epochs]{
\includegraphics[width=0.31\textwidth]{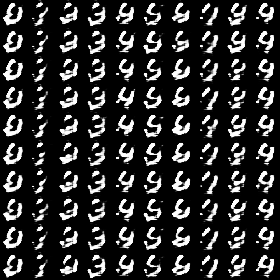}}
\subfigure[\scriptsize AC-GAN 20 epochs]{
\includegraphics[width=0.31\textwidth]{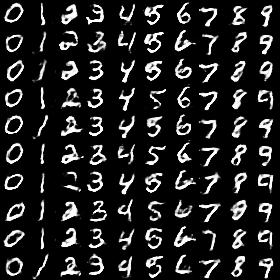}}
\subfigure[\scriptsize AC-GAN 50 epochs]{
\includegraphics[width=0.31\textwidth]{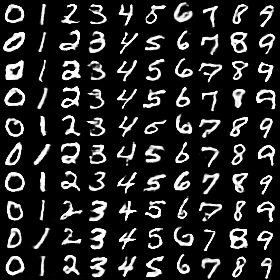}}
\end{minipage}
\begin{minipage}{\columnwidth}
\centering
\subfigure[\scriptsize FC-GAN 10 epochs]{
\includegraphics[width=0.31\textwidth]{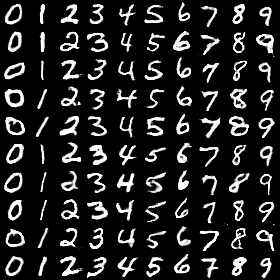}}
\subfigure[\scriptsize FC-GAN 20 epochs]{
\includegraphics[width=0.31\textwidth]{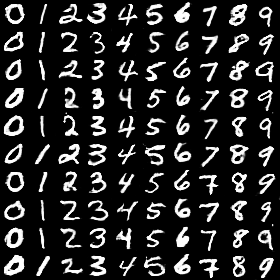}}
\subfigure[\scriptsize FC-GAN 50 epochs]{
\includegraphics[width=0.31\textwidth]{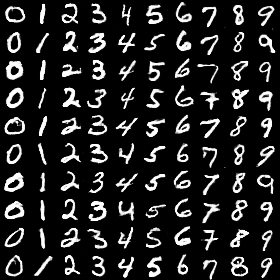}}
\end{minipage}
\caption{The generated images comparison with different epochs on the MNIST dataset.}
\label{fig:generated_mnist}
\end{figure}

The most common metric for generative image models is visual fidelity of generated samples 
\cite{theis2015note}.
Fig.~\ref{fig:generated_mnist} shows the generated images of AC-GAN and FC-GAN after $10$, $20$, and $50$ epochs on the MNIST dataset. Each column is generated by fixing one label class and randomly sampling the latent variable. We observe that the proposed model can achieve promising results with $10$ epochs while AC-GAN needs $20$ or more epochs to achieve comparable results. 
We also observe that the conditioned class label dominates the category of generated images. Varying the latent variable can generate different digit styles. 

\begin{figure}[htb]
\centering
\begin{minipage}{\columnwidth}
\centering
\subfigure[\scriptsize AC-GAN source loss]{
\includegraphics[width=0.48\textwidth]{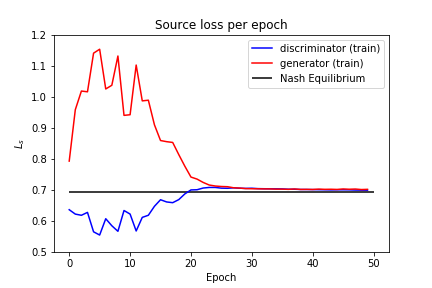}}
\subfigure[\scriptsize FC-GAN source  loss]{
\includegraphics[width=0.48\textwidth]{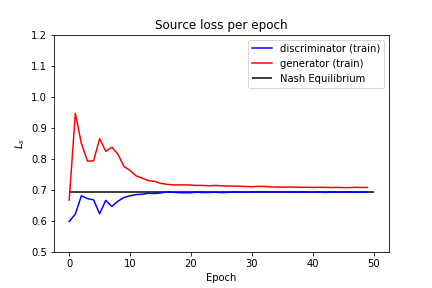}}
\end{minipage}
\caption{The comparison of source loss on MNIST dataset.}
\label{fig:loss_mnist}
\end{figure}

\begin{figure}[htb]
\centering
\begin{minipage}{\columnwidth}
\centering
\subfigure[\scriptsize AC-GAN 10 epochs]{
\includegraphics[width=0.31\textwidth]{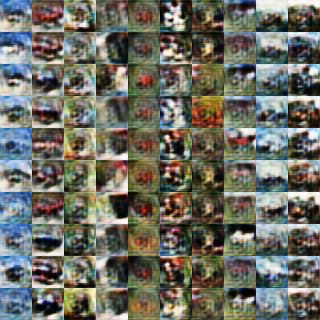}}
\subfigure[\scriptsize AC-GAN 50 epochs]{
\includegraphics[width=0.31\textwidth]{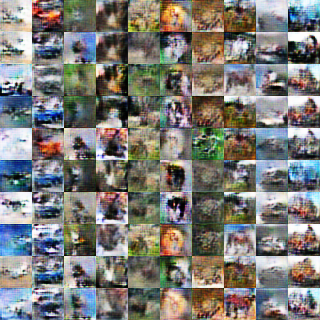}}
\subfigure[\scriptsize AC-GAN 200 epochs]{
\includegraphics[width=0.31\textwidth]{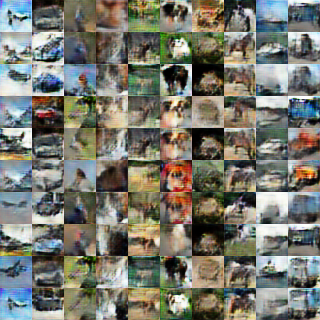}}
\end{minipage}
\begin{minipage}{\columnwidth}
\centering
\subfigure[\scriptsize FC-GAN 20 epochs]{
\includegraphics[width=0.31\textwidth]{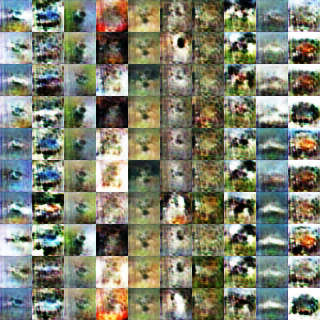}}
\subfigure[\scriptsize FC-GAN 50 epochs]{
\includegraphics[width=0.31\textwidth]{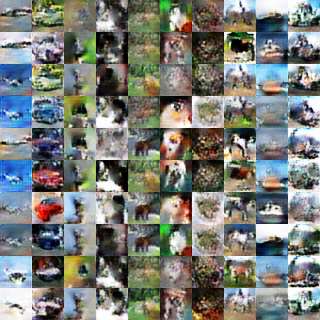}}
\subfigure[\scriptsize FC-GAN 200 epochs]{
\includegraphics[width=0.31\textwidth]{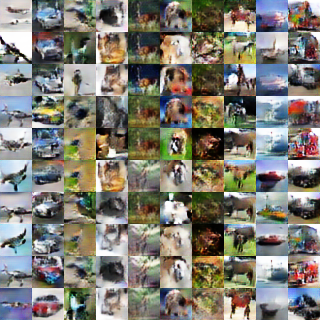}}
\end{minipage}
\caption{The generated images comparison with different epochs on the CIFAR-10 dataset.}
\label{fig:generated_cifar10}
\end{figure}

\begin{figure}[htb]
\centering
\begin{minipage}{\columnwidth}
\centering
\subfigure[\scriptsize AC-GAN source loss]{
\includegraphics[width=0.48\textwidth]{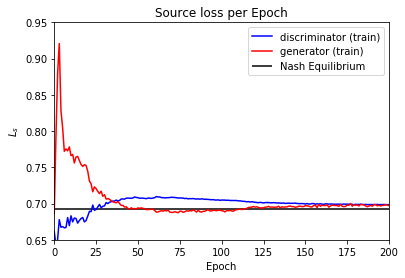}}
\subfigure[\scriptsize FC-GAN source  loss]{
\includegraphics[width=0.48\textwidth]{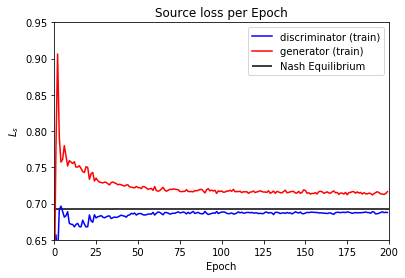}}
\end{minipage}
\caption{The comparison of source loss on CIFAR-10 dataset.}
\label{fig:loss_cifar}
\end{figure}

{The visual fidelity difference between AC-GAN and FC-GAN could also be visualized by the source loss, shown in Fig.~\ref{fig:loss_mnist}. The solution to GANs is the Nash equilibrium which corresponds to $D(x)=D(G(z))=\frac{1}{2}$. Hence, the source losses for the generator and the discriminator are $L_{s}^{G} = -\ln{\frac{1}{2}}=0.693$ and $L_{s}^{D} =-(\ln{\frac{1}{2}}+\ln{\frac{1}{2}})=1.386$, respectively. 
From Fig.~\ref{fig:loss_mnist}, we observe that FC-GAN starts converging at around epoch $13$ while AC-GAN does this at around epoch $20$.} 
We also compare the synthesized images of FC-GAN and AC-GAN with 20, 50 and 200 training epochs and the source loss on CIFAR-10 in Figs.~\ref{fig:generated_cifar10} and \ref{fig:loss_cifar}, respectively. Similar trends are observed here as in the MNIST experiments.
\subsubsection{Parzen Window Estimate}
Exact likelihood of generative adversarial networks is not tractable, and the Parzen window estimate is commonly used as an alternative approach. Our experimental setting follows \cite{goodfellow2014generative}. We first calculate $\sigma$ with a validation set and then fit a Parzen window on randomly generated samples from the generator. Results on the MNIST and CIFAR-10 datasets are reported in Table \ref{tab:parzen-mnist}. 
{Compared with AC-GAN, FC-GAN achieves a slight improvement on MNIST and a significant improvement on CIFAR-10.}
\begin{table}[htb]
\centering
\begin{tabular}{ccc}
\hline
\hline
 & MNIST & CIFAR-10\\
\hline
AC-GAN  & $168.0 \pm 1.4 $ & $581.8 \pm 5.4$\\
FC-GAN  & $175.0 \pm 1.5$    & $646.9 \pm 5.3$ \\
\hline
\end{tabular}
\caption{Parzen window estimates on MNIST and CIFAR-10.}
\label{tab:parzen-mnist}
\vskip -0.1in
\end{table}
\subsubsection{Inception Score}
Inception score was proposed for measuring the performance of generative models which has a high correlation with the quality evaluated by human annotators \cite{odena2016conditional,szegedy2016rethinking}. We show the inception score comparison on the MNIST and CIFAR-10 datasets in Table \ref{tab:incep_score}. The MNIST dataset is much less challenging than CIFAR-10. 
{FC-GAN achieves results comparable to AC-GAN on MNIST, but it does significant improvement on CIFAR-10.}
\begin{table}[htb]
\centering
\begin{tabular}{ccc}
\hline
\hline
 & MNIST & CIFAR-10\\
\hline
AC-GAN & $2.216 \pm 0.04$ & $4.190 \pm 0.08$\\
FC-GAN & $2.238 \pm 0.03$ & $6.360 \pm 0.14$\\
\hline
\end{tabular}
\caption{The inception scores on MNIST and CIFAR-10.}
\label{tab:incep_score}
\vskip -0.1in
\end{table}
\section{Conclusions}
\label{sec:conclusion}
In this paper, we proposed FC-GAN, a fast-converging conditional generative adversarial network. An advanced auxiliary classifier (AC) was introduced for the discriminator, which can distinguish each real class from an extra `fake' class. 
Additionally, the advanced AC also behaves as another real/fake classifier. Experimental results showed that the proposed FC-GAN effectively accelerates the process of differentiation of all classes and helps to generate competitive synthesized images. 
\bibliographystyle{IEEEbib}

\end{document}